%% file: main.tex
\documentclass[letterpaper, 10 pt, conference]{ieeeconf}  %
\IEEEoverridecommandlockouts                              
                                            
\usepackage{cite}
\usepackage{amsmath,amssymb,amsfonts}
\usepackage{algorithmic}
\usepackage{graphicx}
\usepackage{textcomp}
\usepackage{xcolor}
\usepackage{multirow}
\usepackage{comment}
\usepackage{siunitx}

\usepackage[font=footnotesize,labelfont=bf]{caption}
\usepackage[font=footnotesize,labelfont=bf]{subcaption}
\usepackage{balance}
\usepackage[group-separator={,}, per-mode=symbol]{siunitx}

\usepackage{duckuments}
\usepackage{tikzducks}
\usepackage{mathptmx} 
\usepackage[T1]{fontenc}

\makeatletter
\let\NAT@parse\undefined
\makeatother
\usepackage{hyperref}
\usepackage{cleveref}
\hypersetup{
    colorlinks=true,
    linkcolor=blue,
    filecolor=magenta,      
    urlcolor=cyan,
    pdfpagemode=FullScreen,
}

\begin{document}

\title{\LARGE\bf Millimeter Wave Radar: \\From Synthetic Aperture to Probabilistic Mapping}

\author{Jui-Te Huang, Ruoyang Xu, Michael Kaess
\thanks{Jui-Te Huang and Michael Kaess are, Ruoyang Xu was with School of Computer Science, Robotics Institute, Carnegie Mellon University, PA, USA {\tt\footnotesize \{juiteh, ruoyangx, kaess\}@andrew.cmu.edu}}%
\thanks{This work was partially supported by Amazon Lab126 and the U.S. Army Research Office under Contract No. W519TC230031. The content of the information does not reflect the position or the policy of the government, and no official endorsement should be inferred.} 
}


\maketitle

\input{mathabbreviations}
\input{datasetnamedefinitions}
\input{00_abstract.tex}

\input{01_introduction.tex}
\input{02_relatedwork.tex}
\input{03_sar.tex}

\input{04_exps.tex}
\input{05_conclusion}


\bibliographystyle{IEEEtran}
\balance
\bibliography{ref}

\end{document}

%% file: mathabbreviations.tex

\newcommand{\vc}[1]{\boldsymbol{#1}}
\newcommand{\adj}[1]{\frac{d J}{d #1}}
\newcommand{\chain}[2]{\adj{#2} = \adj{#1}\frac{d #1}{d #2}}

\newcommand{\Ac}{\mathcal{A}}
\newcommand{\Bc}{\mathcal{B}}
\newcommand{\Cc}{\mathcal{C}}
\newcommand{\Dc}{\mathcal{D}}
\newcommand{\Ec}{\mathcal{E}}
\newcommand{\Fc}{\mathcal{F}}
\newcommand{\Gc}{\mathcal{G}}
\newcommand{\Hc}{\mathcal{H}}
\newcommand{\Ic}{\mathcal{I}}
\newcommand{\Jc}{\mathcal{J}}
\newcommand{\Kc}{\mathcal{K}}
\newcommand{\Lc}{\mathcal{L}}
\newcommand{\Mc}{\mathcal{M}}
\newcommand{\Nc}{\mathcal{N}}
\newcommand{\Oc}{\mathcal{O}}
\newcommand{\Pc}{\mathcal{P}}
\newcommand{\Qc}{\mathcal{Q}}
\newcommand{\Rc}{\mathcal{R}}
\newcommand{\Sc}{\mathcal{S}}
\newcommand{\Tc}{\mathcal{T}}
\newcommand{\Uc}{\mathcal{U}}
\newcommand{\Vc}{\mathcal{V}}
\newcommand{\Wc}{\mathcal{W}}
\newcommand{\Xc}{\mathcal{X}}
\newcommand{\Yc}{\mathcal{Y}}
\newcommand{\Zc}{\mathcal{Z}}

\newcommand{\Ab}{\mathbb{A}}
\newcommand{\Bb}{\mathbb{B}}
\newcommand{\Cb}{\mathbb{C}}
\newcommand{\Db}{\mathbb{D}}
\newcommand{\Eb}{\mathbb{E}}
\newcommand{\Fb}{\mathbb{F}}
\newcommand{\Gb}{\mathbb{G}}
\newcommand{\Hb}{\mathbb{H}}
\newcommand{\Ib}{\mathbb{I}}
\newcommand{\Jb}{\mathbb{J}}
\newcommand{\Kb}{\mathbb{K}}
\newcommand{\Lb}{\mathbb{L}}
\newcommand{\Mb}{\mathbb{M}}
\newcommand{\Nb}{\mathbb{N}}
\newcommand{\Ob}{\mathbb{O}}
\newcommand{\Pb}{\mathbb{P}}
\newcommand{\Qb}{\mathbb{Q}}
\newcommand{\Rb}{\mathbb{R}}
\newcommand{\Sb}{\mathbb{S}}
\newcommand{\Tb}{\mathbb{T}}
\newcommand{\Ub}{\mathbb{U}}
\newcommand{\Vb}{\mathbb{V}}
\newcommand{\Wb}{\mathbb{W}}
\newcommand{\Xb}{\mathbb{X}}
\newcommand{\Yb}{\mathbb{Y}}
\newcommand{\Zb}{\mathbb{Z}}

\newcommand{\av}{\mathbf{a}}
\newcommand{\bv}{\mathbf{b}}
\newcommand{\cv}{\mathbf{c}}
\newcommand{\dv}{\mathbf{d}}
\newcommand{\ev}{\mathbf{e}}
\newcommand{\fv}{\mathbf{f}}
\newcommand{\gv}{\mathbf{g}}
\newcommand{\hv}{\mathbf{h}}
\newcommand{\iv}{\mathbf{i}}
\newcommand{\jv}{\mathbf{j}}
\newcommand{\kv}{\mathbf{k}}
\newcommand{\lv}{\mathbf{l}}
\newcommand{\mv}{\mathbf{m}}
\newcommand{\nv}{\mathbf{n}}
\newcommand{\ov}{\mathbf{o}}
\newcommand{\pv}{\mathbf{p}}
\newcommand{\qv}{\mathbf{q}}
\newcommand{\rv}{\mathbf{r}}
\newcommand{\sv}{\mathbf{s}}
\newcommand{\tv}{\mathbf{t}}
\newcommand{\uv}{\mathbf{u}}
\newcommand{\vv}{\mathbf{v}}
\newcommand{\wv}{\mathbf{w}}
\newcommand{\xv}{\mathbf{x}}
\newcommand{\yv}{\mathbf{y}}
\newcommand{\zv}{\mathbf{z}}

\newcommand{\Av}{\mathbf{A}}
\newcommand{\Bv}{\mathbf{B}}
\newcommand{\Cv}{\mathbf{C}}
\newcommand{\Dv}{\mathbf{D}}
\newcommand{\Ev}{\mathbf{E}}
\newcommand{\Fv}{\mathbf{F}}
\newcommand{\Gv}{\mathbf{G}}
\newcommand{\Hv}{\mathbf{H}}
\newcommand{\Iv}{\mathbf{I}}
\newcommand{\Jv}{\mathbf{J}}
\newcommand{\Kv}{\mathbf{K}}
\newcommand{\Lv}{\mathbf{L}}
\newcommand{\Mv}{\mathbf{M}}
\newcommand{\Nv}{\mathbf{N}}
\newcommand{\Ov}{\mathbf{O}}
\newcommand{\Pv}{\mathbf{P}}
\newcommand{\Qv}{\mathbf{Q}}
\newcommand{\Rv}{\mathbf{R}}
\newcommand{\Sv}{\mathbf{S}}
\newcommand{\Tv}{\mathbf{T}}
\newcommand{\Uv}{\mathbf{U}}
\newcommand{\Vv}{\mathbf{V}}
\newcommand{\Wv}{\mathbf{W}}
\newcommand{\Xv}{\mathbf{X}}
\newcommand{\Yv}{\mathbf{Y}}
\newcommand{\Zv}{\mathbf{Z}}

\newcommand{\alphav     }{\boldsymbol \alpha     }
\newcommand{\betav      }{\boldsymbol \beta      }
\newcommand{\gammav     }{\boldsymbol \gamma     }
\newcommand{\deltav     }{\boldsymbol \delta     }
\newcommand{\epsilonv   }{\boldsymbol \epsilon   }
\newcommand{\varepsilonv}{\boldsymbol \varepsilon}
\newcommand{\zetav      }{\boldsymbol \zeta      }
\newcommand{\etav       }{\boldsymbol \eta       }
\newcommand{\thetav     }{\boldsymbol \theta     }
\newcommand{\varthetav  }{\boldsymbol \vartheta  }
\newcommand{\iotav      }{\boldsymbol \iota      }
\newcommand{\kappav     }{\boldsymbol \kappa     }
\newcommand{\varkappav  }{\boldsymbol \varkappa  }
\newcommand{\lambdav    }{\boldsymbol \lambda    }
\newcommand{\muv        }{\boldsymbol \mu        }
\newcommand{\nuv        }{\boldsymbol \nu        }
\newcommand{\xiv        }{\boldsymbol \xi        }
\newcommand{\omicronv   }{\boldsymbol \omicron   }
\newcommand{\piv        }{\boldsymbol \pi        }
\newcommand{\varpiv     }{\boldsymbol \varpi     }
\newcommand{\rhov       }{\boldsymbol \rho       }
\newcommand{\varrhov    }{\boldsymbol \varrho    }
\newcommand{\sigmav     }{\boldsymbol \sigma     }
\newcommand{\varsigmav  }{\boldsymbol \varsigma  }
\newcommand{\tauv       }{\boldsymbol \tau       }
\newcommand{\upsilonv   }{\boldsymbol \upsilon   }
\newcommand{\phiv       }{\boldsymbol \phi       }
\newcommand{\varphiv    }{\boldsymbol \varphi    }
\newcommand{\chiv       }{\boldsymbol \chi       }
\newcommand{\psiv       }{\boldsymbol \psi       }
\newcommand{\omegav     }{\boldsymbol \omega     }

\newcommand{\Gammav     }{\boldsymbol \Gamma     }
\newcommand{\Deltav     }{\boldsymbol \Delta     }
\newcommand{\Thetav     }{\boldsymbol \Theta     }
\newcommand{\Lambdav    }{\boldsymbol \Lambda    }
\newcommand{\Xiv        }{\boldsymbol \Xi        }
\newcommand{\Piv        }{\boldsymbol \Pi        }
\newcommand{\Sigmav     }{\boldsymbol \Sigma     }
\newcommand{\Upsilonv   }{\boldsymbol \Upsilon   }
\newcommand{\Phiv       }{\boldsymbol \Phi       }
\newcommand{\Psiv       }{\boldsymbol \Psi       }
\newcommand{\Omegav     }{\boldsymbol \Omega     }

%% file: datasetnamedefinitions.tex
\newcommand{\nsh}{\textbf{north}}
\newcommand{\nshshort}{\textbf{north\_short}}
\newcommand{\nsha}{\textbf{north\_a}}
\newcommand{\nshb}{\textbf{north\_b}}
\newcommand{\wean}{\textbf{west\_2f}}

%% file: 00_abstract.tex
\begin{abstract}
Robust probabilistic mapping is essential for autonomous robotic systems operating in challenging environments. While traditional sensors fail in adverse conditions such as smoke and fog, millimeter wave (mmWave) radar sensors offer reliable sensing in such conditions. However, creating accurate probabilistic maps from radar data presents significant challenges due to the inherently sparse and noisy characteristics of radio wave measurements and signal processing steps. In an attempt to address these issues, we establish a complete pipeline from raw radar signals to probabilistic occupancy maps, incorporating Synthetic Aperture Radar processing followed by a probabilistic modeling step. We conduct extensive validation across indoor environments, comparing our approach against different signal processing and probabilistic modeling approaches. We also evaluate mapping quality through downstream path planning performance analysis. Furthermore, we investigate the impact of key parameters and antenna array configuration on mapping performance. The experimental results demonstrate both the effectiveness and limitations of SAR-based probabilistic mapping for real-world robotic deployment. To facilitate future research and broader adoption, we contribute an open-source cascaded mmWave radar dataset with an accompanying GPU-accelerated signal processing pipeline available at \href{https://github.com/rpl-cmu/rpm}{https://github.com/rpl-cmu/rpm}.

\end{abstract}


%% file: 01_introduction.tex
\section{Introduction}
Probabilistic mapping is a fundamental building block in the robot autonomy stack, enabling downstream tasks such as exploration, planning, and control. Prior research has demonstrated successful and efficient probabilistic mapping using time-of-flight sensors like LiDARs and depth cameras\cite{hornung13auro}. However, these sensors experience significant performance degradation in adverse environments such as smoke and fog, where airborne particles obstruct light rays before they reach actual objects.

On the other hand, millimeter wave (mmWave) radar sensors transmit and receive electromagnetic waves with millimeter-level wavelengths, providing robust sensing capabilities in adverse environmental conditions without relying on external lighting or heating. Recent developments in next-generation mmWave radars utilize novel modulation schemes and multiple antenna packaging to achieve higher spatial resolution. Consequently, radar sensors have attracted significant attention from researchers applying this technology for object detection \cite{wang2021rodnet, wang2021rod2021, wang2024vision}, navigation \cite{huang2021cross}, state estimation \cite{park20213d, doer2021x, kramer2022coloradar, lu2020milliego, huang2024multi}, novel view synthesis \cite{huang2024dart} and mapping \cite{huang2024dart, borts2024radar, Xu22iros, zhang2024towards, ding2024radarocc, mopidevi2024rmap, prabhakara2023high, kung2025radarsplat}.

While research on radar-based state estimation is achieving performance comparable to LiDAR and camera-based methods, radar-based mapping remains challenging due to the sparse and noisy nature of processed mmWave radar data. Several studies have demonstrated promising results by training neural networks to generate dense scene geometry under the supervision of LiDAR data\cite{Huang_2025_ICCV}. However, these approaches struggle to generalize to novel environments and radar sensors from different manufacturers, as variations in signal modulation parameters, antenna layouts, and radiation patterns result in different fields of view and noise characteristics.

\input{figs/teaser}


Unlike machine learning approaches, Synthetic Aperture Radar (SAR) is a technique for mapping that leverages radio wave coherence and has been used to map the Earth's surface from satellites hundreds of kilometers above. Researchers have applied this technique to modern mmWave radar for concealed object detection \cite{yanik2020development}, automobile mapping \cite{grebner2023probabilistic,grebner2022radar}, and mobile robot mapping \cite{ritterbusch2024indoor,ritterbusch2024rio,ritterbusch2024simultaneous}. Despite these recent developments, adapting SAR techniques for probabilistic mapping is rarely discussed and requires thorough evaluation before application in real-world robotic systems. 

In this work, we adapted synthetic aperture radar mapping techniques for robotic mapping tasks. We introduce an approach to convert SAR maps into probabilistic maps by accounting for occlusions and object probability distributions. 
To our knowledge, no prior work has demonstrated and evaluated a complete pipeline from raw automotive mmWave signals to probabilistic occupancy maps suitable for downstream planning tasks.
Our approach is validated across multiple indoor scenarios and benchmarked against baseline methods. We also provide a comprehensive analysis of the method's limitations through systematic experiments. Our key contributions include:

\begin{itemize}
    \item \textbf{From Radio Signal to Probabilistic Map:}
    We introduce a complete pipeline for mmWave radar mapping, from radar modulation scheme to synthetic aperture radar processing to probabilistic occupancy maps.
    We evaluate our pipeline against two common mmWave radar signal processing approaches using LiDAR-generated ground truth maps on map fidelity and usefulness for downstream path planning performance. We additionally perform ablation studies on key parameters and antenna configurations.

       
    \item \textbf{Open Source Dataset and Code:} 
        Our cascaded mmWave radar signal dataset with GPU-accelerated signal processing pipeline is publicly available online to facilitate reproducibility and further research. 
    
\end{itemize}
\vspace{-3mm}

%% file: figs/teaser.tex
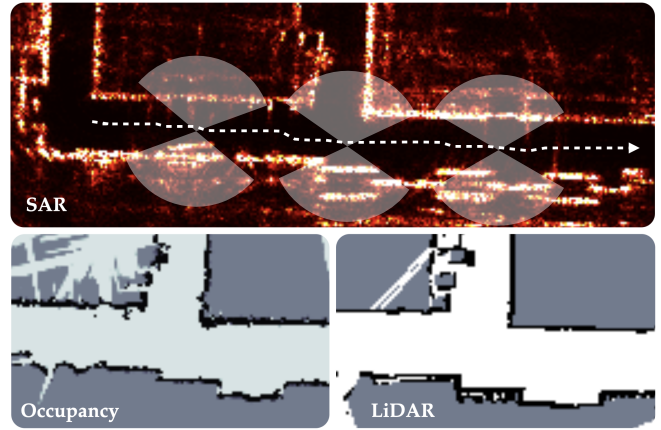
\begin{figure}
    \centering
\begin{tikzpicture}
    \node[outer sep=0pt, inner sep=0pt, anchor=north west] (base_image) at (0, 0) {
        \includegraphics[width=\linewidth]{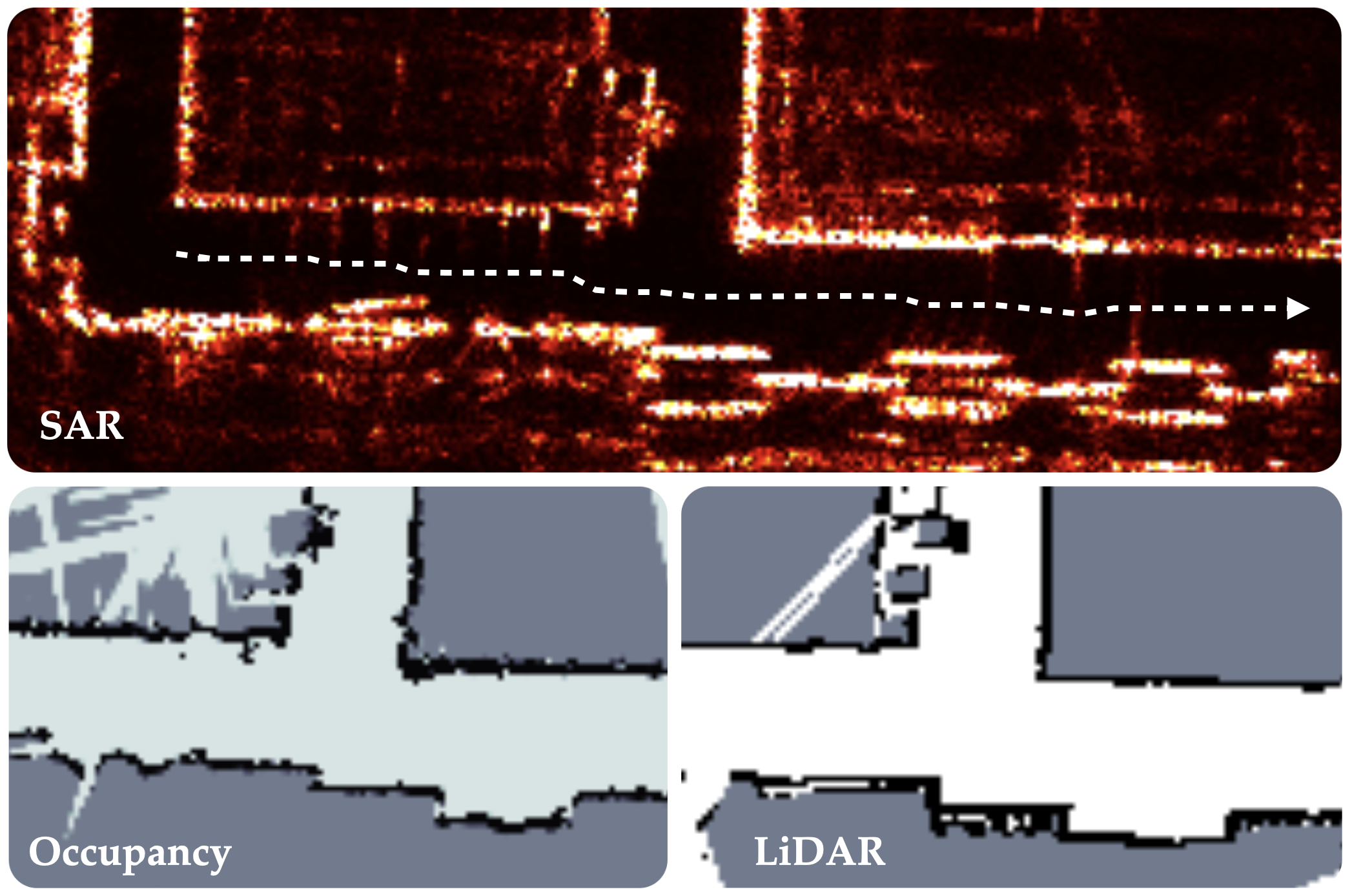}
    };

    \fill[fill=white, fill opacity=0.4, shift={(6.5, -2)}] (0,0) -- (30:1cm) arc (30:150:1cm) -- cycle;
    \fill[fill=white, fill opacity=0.4, shift={(6.5, -2)}] (0,0) -- (210:1cm) arc (210:330:1cm) -- cycle;

    \fill[fill=white, fill opacity=0.4, shift={(2.5, -1.75)}] (0,0) -- (20:1cm) arc (20:140:1cm) -- cycle;
    \fill[fill=white, fill opacity=0.4, shift={(2.5, -1.75)}] (0,0) -- (200:1cm) arc (200:320:1cm) -- cycle;

    \fill[fill=white, fill opacity=0.4, shift={(4.5, -1.95)}] (0,0) -- (25:1cm) arc (25:145:1cm) -- cycle;
    \fill[fill=white, fill opacity=0.4, shift={(4.5, -1.95)}] (0,0) -- (205:1cm) arc (205:325:1cm) -- cycle;


\end{tikzpicture}

\caption{A demonstration of our proposed method using two cascade mmWave radar boards to create a LiDAR-like occupancy map (bottom right). As a vehicle moves through the environment, onboard radar sensors create a synthetic aperture to map the surroundings (Top). A probability modeling method is presented to create the occupancy map (bottom left).}
    \label{fig:teaser}
    \vspace{-5mm}
\end{figure}

%% file: 02_relatedwork.tex
\section{Related Work}

\subsection{Probability Mapping with Radar}
Recent developments in industrial and automotive mmWave radar sensors have enabled research for mapping using radio waves. A popular strategy is to first detect targets using the Constant False Alarm Rate (CFAR) algorithm, then carefully model the probability of targets for occupancy mapping \cite{werber2015automotive, degerman20163d, prophet2018adaptions, kramer2020radar}. This highlights a major difference compared to the well-established probablistic mapping method in \cite{hornung13auro}, where the strategy attempts to remove the effect of occlusion and preserve as many high probability targets as possible. Another idea proposed by \cite{weishaupt2022precfar} uses the complete radar cube data with a strategy for finding the static objects' Doppler bins to create a range-azimuth amplitude heatmap for mapping. However, the map was created using a simple averaging operation of scattering power in the Cartesian space rather than a probability map.

\subsection{Synthetic Aperture Radar}
While the above methods employed detected targets or amplitude heatmaps for mapping, researchers have also investigated synthetic aperture radar for mapping both outdoor road scenes \cite{iqbal2021realistic, grebner2022radar} and indoor environments with mobile robots \cite{ritterbusch2024indoor, ritterbusch2024rio, ritterbusch2024simultaneous}. In the majority of these studies, maps were represented in terms of scattering power. Grebner et al. \cite{grebner2023probabilistic} demonstrated an approach for deriving probability maps from SAR processing by separately modeling amplitude and phase distributions, then strategically combining these models to update the final probability map. This approach, however, demands highly accurate probability distribution parameters, as inaccuracies can cause probabilities to rapidly converge to either 0 or 1 when processing tens of thousands of radio measurements. In this work, we introduce and evaluate an alternative methodology for transforming synthetic aperture radar maps into occupancy maps. 

\subsection{Learning Methods}
Many works have adapted supervised learning solutions to generate high-quality point clouds or maps to improve radar mapping quality. In \cite{lu2020see}, Lu et al. trained neural networks to predict 2D grid map patches using conditional GANs \cite{wang2018high}, while \cite{mopidevi2024rmap} output 3D occupancy map patches. Xu et al. \cite{Xu22iros} trained neural nets to output depth images, while \cite{zhang2024towards} and \cite{prabhakara2023high} output high-resolution range-azimuth images. Lai et al. \cite{panoradar} further placed the mmWave radar on a rotating platform, using beamforming techniques and collected frames to create a cylindrical array imaging, then employed a neural network to output LiDAR-quality depth images and semantic segmentation. Instead of supervised learning, \cite{huang2024dart, borts2024radar} used multi-views of radar data to optimize neural scene representation and created maps as a byproduct. However, these maps are not evaluated with downstream tasks and are only served for visualization. While the above machine learning approaches showed promising results, neural networks can be difficult to generalize to different radio modulation parameters and antenna designs. In this work, we discuss and demonstrate direct probability mapping methods using radio wave coherence and probability modeling without the need for a supervised machine learning approach.

\vspace{-4mm}

%% file: 03_sar.tex
\section{Methodology}
In this section, we introduce our approach for 2-dimensional probability mapping using mmWave radar. We assume that the poses of all radar antennas involved in transmitting and receiving signals are known. Our objective is to determine the probability of occupancy for each cell in a 2D map. Radar signal readings and estimated poses arrive sequentially, allowing us to incrementally build and update the map. In the following subsections, we first provide background on mmWave radio signals and synthetic aperture radar (SAR) mapping. We then present our method for converting SAR maps into probability maps.
\subsection{Frequency Modulated Continuous Wave}
Modern mmWave radar sensors for automotive and industrial applications use a modulation scheme called Frequency Modulated Continuous Wave (FMCW). A type of FMCW can transmit chirp signals with a frequency growing linearly in time. The transmitted signal $T_X(t)$ can be written as complex sinusoids with frequency slope $\mu$ and starting frequency $f$. And the received signal return from a single object at a distance $r$ is simply a time-delayed transmit signal with EM wave transmission time $\tau_0 = 2r / c$ :
\begin{align}
    T_X(t) &= \exp(j(\pi\mu t^2+2\pi ft)) \\
    R_X(t) &= \exp(j(\pi\mu (t-\tau_0)^2+2\pi f(t-\tau_0)))
\end{align}
On the mmWave radar sensors, we recorded the intermediate frequency (IF) signal $s(t)$, which is the correlation between the transmit and received signal.
\begin{align}
    s(t) &= T_X(t) \otimes R_X(t) \\
    &= \exp\left(j\left(2\pi\mu\frac{2r}{c}t+2\pi f\frac{2r}{c}-\pi\mu\frac{4r^2}{c^2}\right)\right) \\
    s(t) &\approx \exp\left(j\frac{4\pi r}{c}(f+\mu t)\right)
\end{align}

 Both the frequency and the phase of $s(t)$ are functions of distance to the object. In a real-world scenario, there can be multiple reflected signals from different distances, and the received signal is the superposition of all the reflected signals. Finally, each IF signal is sampled through an analog-to-digital converter and recorded digitally.

 In each frame of radar data, a set number of chirps are sent through each transmit antenna sequentially and received by all receive antennas. We can rearrange a frame of signals to a radar cube of complex numbers with size $(samples, chirps, Tx, Rx)$ for downstream signal processing or process every IF signal individually for synthetic aperture radar mapping.

\begin{figure*}[t]
    \centering
    \includegraphics[width=.8\linewidth]{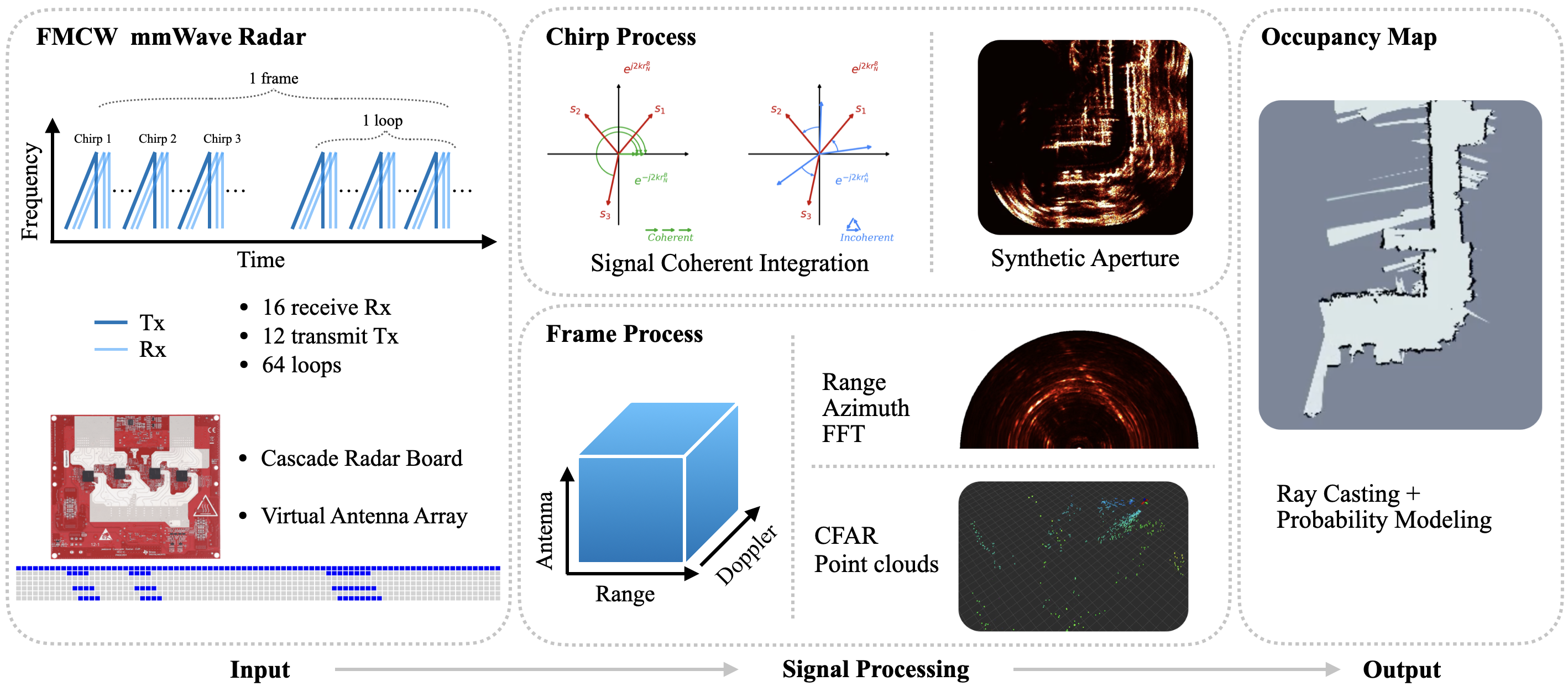}
    \caption{Radar mapping procedures from raw signal to a probability map. In this work, we demonstrate using different signal processing steps to generate a probability map. We can jointly process multiple chirps of one radar frame or process each chirp separately to create a synthetic aperture for mapping.}
    \label{fig:mapping_steps}

    \vspace{-5mm}
\end{figure*}
\vspace{-5mm}
\subsection{Synthetic Aperture Radar Mapping}
To map the environment using radio waves, we employ the back-projection (BP) algorithm \cite{ulander2003synthetic, curlander1991synthetic, frey2009focusing}, which is suitable for handling non-linear sensor paths. For a point of interest $\Vec{q}$ on the map, the basic principle of BP is to coherently integrate all measurements of the radar signal samples $s(\Vec{p}, m)$ $s_{\Vec{p}}[m]$ collected at position $\Vec{p}$ with sampling interval $t_s$ and total $M$ samples for each chirp. Therefore, we maintain a 2-dimensional grid map $\Mv(\Vec{q})$ where each grid cell stores a complex number. 
\begin{align}
    \Mv(\Vec{q}) &= \sum_{\Vec{p}\in\Pc}\sum_{m=0}^{M-1} s(\Vec{p},m)\exp\left(-j\frac{4\pi d}{c}(f+\mu mt_s)\right) \\
    d&=\left(|\Vec{p}_{tx}-\Vec{q}|+|\Vec{p}_{rx}-\Vec{q}|\right)/2
\end{align}

For every signal measured, we need to unrotate the phase according to the expected total wave transmission distance. Notice that the transmit distance $2d$ is from the position of the transmit antenna $\Vec{p}_{tx}$ to the grid position $\Vec{q}$, then back to the received antenna $\Vec{p}_{rx}$. This phase correction procedure ensures that the true targets on the map add up constructively since their phases are more aligned. And for the grid positions with no actual target, the complex numbers tend to cancel out each other since their corrected phase will not be aligned. After the radar moves along a path to create a synthetic aperture, grid positions containing obstacles exhibit higher amplitude compared to empty space.

While the BP algorithm can map the environment using signals that are not collected from a linear path, it runs slowly since we need to iterate through all time samples $T$ and signals collected within the field of view. We speed up the process by separating the summation related to time samples.
\begin{align}
    &M(\Vec{q}) \notag\\
    &= \sum_{\Vec{p}\in\Pc}\exp\left(-j\frac{4\pi}{c}fd\right)\sum_{m=0}^{M-1}s(\Vec{p},m)\exp\left(-j\frac{4\pi}{c}\mu mt_sd\right) \\
    &= \sum_{\Vec{p}\in\Pc}\exp\left(-j\frac{4\pi}{c}fd\right)\sum_{m=0}^{M-1}s(\Vec{p},m)\exp\left(-j2\pi\frac{2BTd}{cT}\frac{m}{M}\right)
\end{align}
where $T=Mt_s$ is the total sampling time and $B=\mu T$ is the chirp bandwidth. This is equivalent to performing Fast Fourier Transform (FFT) on the chirp signal and then picking the complex value at the corresponding bin.
\begin{align}
    S(\Vec{p},k) &= \sum_{m=0}^{M-1}s(\Vec{p},m)\exp\left(-j2\pi k\frac{m}{M}\right) \\
    \Mv(\Vec{q}) &= \sum_{\Vec{p}\in\Pc}\exp\left(-j\frac{4\pi}{c}fd\right)S\left(\Vec{p},\frac{2B}{c}d\right)
    \label{eq: bp}
\end{align}
Since k is a discrete number, we interpolate $S$ to find the corresponding complex number to add to $\Mv(\Vec{q})$. We can reuse $S$ without recomputing the inner summation. 

To incrementally map the environment, for each sensor position and received signal $z_t$, we identify the corresponding grid cells on map $\Mv$ within the radar's field of view to update. We then calculate the value to add to each grid cell according to Equation~\ref{eq: bp}. In addition to recording complex values, we also track the $\Mv_n$ number of times each grid cell has been updated. Furthermore, we adjust the projected signal intensity according to the antenna radiation pattern, which was simulated using MATLAB based on our radar sensor antenna design.

\subsection{Occupancy Mapping}

While mapping using synthetic aperture radar data, we are updating a grid map $\Mv$ with complex values where higher amplitude suggests the existence of objects on the grid. However, these complex values are unbounded and may increase with more updates since our robot path is highly non-linear, resulting in some grid cells being updated more frequently than others. To create a probability map using radar, we model the probability of obstacle presence on the complex value map $\Mv$. The Rayleigh distribution~\cite{kuruoglu2004modeling} provides a well-established statistical framework for modeling this probability. We model the probability using the normalized amplitude of each grid cell by the CDF of the Rayleigh distribution. The scale parameter $\sigma_a$ is set according to our radar sensors' thermal noise.
\begin{align}
    \Rv &= 1-\exp\left( \frac{-\left(|\Mv|/\Mv_n\right)^2}{2\sigma_a^2} \right)
\end{align}

Then, to differentiate between free space and unknown space, we perform ray casting on the probability map $\Rv$ within the field of view of the current pose. For each ray $r$, we model the transmittance $T$ along the range according to the probability values on $\Rv$. This step also helps us remove the artifacts created by the multi-path signal.
\begin{align}
    T(n|z_t) &= \prod_{n\in r} \left(1-\Rv(n|z_t) \right) 
    \label{eq: transmittance}    
\end{align}

These SAR map readings $\Rv$ on each ray are then integrated incrementally using a strategy modified from occupancy grid mapping~\cite{hornung13auro}. For Occupancy mapping, the probability of a cell on map $\Pv(n|z_{1:t})$ given the sensor measurements $z_{1:t}$ is estimated as:
\begin{align}
    \Pv(n|z_{1:t}) &= \left[1+\frac{1-\Pv(n|z_{t})}{\Pv(n|z_{t})}\frac{1-\Pv(n|z_{1:t-1})}{\Pv(n|z_{1:t-1})} \right]^{-1}
\end{align}

Using the log-odds notation, we can safely handle small probabilities without losing numerical stability.
\begin{align}
    \Lv(n|z_{1:t}) &= \Lv(n|z_{1:t-1}) + \Lv(n|z_t) \\
    \Lv(n) &= \log \left[\frac{\Pv(n)}{(1-\Pv(n))} \right]
\end{align}


We then maintain and update the map $\Lv$, which stores log-odds values to account for transmittance and probability. This step is performed every time after the synthetic aperture radar map $\Mv$ is updated by new signal measurements and pose pairs $z_t$.
\begin{align}
    \Lv(n) &= T(n)\log\left(\frac{\Rv(n)}{1-\Rv(n)}\right) 
    \label{eq: log-odds}
\end{align}

Similar to~\cite{hornung13auro}, we have set the lower and upper bounds on the SAR map readings $\Rv$ and log-odds map $\Lv$ to avoid saturation, ensuring the map can adapt to changes quickly. Finally, the log-odds values $\Lv$ is converted back to the probability $\Pv$ for downstream applications and visualization.

To ensure the SAR map $\Rv$ image is properly focused before updating the occupancy map, we delay the Occupancy Mapping step by two frames. While the field of view for SAR mapping encompasses the entire front field of view of the radar sensors, we only use a 30-degree field of view for updating map $\Lv$ to ensure we do not update the occupancy map with unfocused SAR map regions. This helps keep the area behind the first obstacles in the line of sight as an unknown region, rather than incorrectly modeling it as empty space due to the not fully focused SAR map $\Rv$.

%% file: 04_exps.tex
\section{Experiments}

We demonstrate and evaluate our proposed method for mmWave radar mapping in various indoor scenarios using a ground robot equipped with two off-the-shelf cascade mmWave radar sensors on opposite sides. In the following section, we will first introduce our sensor platform setup, then evaluate the fidelity of our proposed method with reference to LiDAR maps, as well as the applicability of our proposed method for robot path planning. Finally, we conduct an ablation study on the impact of the number of antennas and pose drift to map quality.




\subsection{Data Collection} 
Our sensor platform is shown in Figure~\ref{fig:sensor-platform-img}. It consists of:
\begin{itemize}
    \item Two Texas Instrument MMWCAS-RF-EVM mmWave cascaded imaging radar at 10Hz,
    \item Two PointGrey BlackFly GigE machine cameras at 20Hz,
    \item One Velodyne VLP-16 LiDAR at 10Hz,
    \item One Epson G-364 IMU at 200Hz.
\end{itemize}
The sensor platform is mounted on an RC car with Ackermann steering, and the sensors are synchronized through a real-time clock on a Teensy board. Both of our radars use the same set of signal configurations as outlined in Table~\ref{tbl:radar-signal-config}. We deemed this repetition acceptable as the sensors face away from each other and are on two opposite sides of the sensor rig. With these radar signals
settings, we achieve a maximum range of \SI{15}{\meter}, range resolution of \SI{6}{\centi\meter}.

We collected data for indoor scenarios from campus buildings. The collected sequences consist of corridors with varying degrees of width and clutter, and small patches of open workspaces with chairs and desks, as illustrated in Figure~\ref{fig:sequence-example-pic}. We purposefully avoided classrooms and offices because the navigable space is typically too tight for our robot to have a meaningful amount of movement for SAR to map the room properly.

\input{figs/sensor_platform}
\input{figs/signal_config}

\subsection{Ground Truth and Baselines}
\subsubsection{Ground Truth Pose}
To test our mapping algorithm independently of pose estimation errors, we obtained pseudo ground truth poses using the LiDAR-inertial odometry algorithm FAST-LIO~\cite{xu2021fast} or the stereo visual-inertial odometry algorithm openVINS~\cite{Geneva2020ICRA} when the LiDAR odometry failed in a few long corridor scenes. Since radar chirp signals are transmitted at a higher frequency than our odometry source, we interpolated the pseudo ground truth poses to determine the position of each transmit and receive antenna for every received intermediate frequency (IF) signal.

\subsubsection{Ground Truth Map}
The ground truth map was generated using occupancy mapping~\cite{hornung13auro} with undistorted LiDAR point clouds. To ensure fair comparison across all methods, the resolution of each grid cell was set to 0.1 meters for both the tested algorithms and the ground truth map.

\subsubsection{Baseline Methods}
We compare our proposed method against two other approaches that involve different signal processing procedures to create a map.

\textbf{CFAR:} The Constant False Alarm Rate (CFAR) is a common algorithm to get point clouds from mmWave radar signal. Firstly, we run a 2D FFT and FFT shift on each frame of signal data along the sample dimension and the chirp dimension to obtain Range-Doppler images. We then run Cell Averaging Smallest of CFAR (CASO-CFAR) to detect objects with strong reflection. Then we can rearrange the antenna phasors to the virtual antenna position as shown in Fig.~\ref{fig:mapping_steps} and run another 2D FFT to get azimuth and elevation angles. Finally, the 3D information is converted from polar coordinates to Cartesian coordinates with Doppler velocity and signal-to-noise ratio. We ran this algorithm provided by the TI development tool\cite{texas2020imaging} to get the radar point clouds. 

With the radar points, we first filtered out noisy points with Doppler velocity that are not aligned with the estimated linear velocity, similar to many state estimation algorithms\cite{huang2024multi}. Then the radar points within the field of view and above the estimated floor height were used by OctoMap\cite{hornung13auro} to create 2D occupancy maps. We set the maximum mapping distance to 10 meters and the grid resolution to 0.1 meters, the same as our proposed algorithm.

\textbf{Range-Azimuth:} The Range-Doppler-Angle FFT is a common signal processing step to obtain a 3D radar cube, which is used by many radar-based object detection algorithms\cite{wang2021rodnet}. We can take the mean on the Doppler axis to create a Range-Azimuth image that roughly outlines the environment. To process our signals to a Range-Azimuth image, we first apply range-FFT on the sample dimension and then rearrange the antenna phasors to the virtual antenna position. Then we run the angle-FFT, followed by FFT shift to obtain a Range-Azimuth-Doppler cube. Finally, we drop the last axis by taking the mean of the amplitude to obtain our Range-Azimuth image.

To update the occupancy map from this Range-Azimuth image, we applied the same strategy as our proposed method. We first model the probability of our Range-Azimuth amplitude using the Rayleigh distribution, then for each range bin, the transmittance and log-odds are calculated same way as shown in equations \ref{eq: transmittance} and \ref{eq: log-odds}. Finally, the occupancy map was updated by the log-odds value according to the sensor pose and the coordinate of each Range-Azimuth pixels.

\subsection{Parameter Details}

Our mapping pipeline involves parameters spanning synthetic aperture, probabilistic modeling, and occupancy mapping. The key parameters are summarized in Table~\ref{table: parameter}.                               

\input{figs/parameters_table}

Among these, the Rayleigh distribution scale parameter $\sigma_a$ is the least intuitive to set. It reflects the noise floor of the radar sensor and should be calibrated to the specific sensor in use. We perform a parameter search over $\sigma_a$ and evaluate the F-score against the ground-truth occupancy map across multiple sequences, as shown in Figure~\ref{fig: sigmas}. The best average performance is achieved at $\sigma_a \approx 0.15$.

\input{figs/sigma_ablation}
\input{figs/sequence_example_pic}
\input{figs/map_quant_table}

\subsection{Map Evaluations}

\subsubsection{Quantitative Evaluation}
To evaluate our radar occupancy map quality, we first evaluate the cells that are considered obstacles. We took the cells that have a probability greater than 0.6 and transformed them into 2D points. We assessed these points against ground-truth points from the LiDAR occupancy map and computed the Chamfer Distance (CD), Hausdorff Distance (HD), and F-score. The distance threshold to compute precision and recall for the F-score was set at 0.2 meters. The results are shown in TABLE~\ref{table: point eval}. Our proposed method has the best CD score across 10 different sequences. For some second-best performance of HD and F-score was caused by outlier points, which are not close to the real mapping area. The baseline algorithm CFAR generates the noisiest points, even with the Doppler velocity outlier removal, it is not ideal for updating the occupancy map. While the Range-Azimuth mapping approach suffers from getting different amplitude responses for the same area from different viewing angles due to the surface reflection of electromagnetic waves. Without any coherent integration, the occupied cells will easily be eliminated. 


\subsubsection{Qualitative Evaluation} 

Examples of the mapping environment, LiDAR built occupancy map, and radar built occupancy map with different baseline approaches and our proposed approach are in Figure~\ref{fig:sequence-example-pic}. Our SAR-based approach clearly outlined the geometry of the corridor compared to baseline methods, where environment geometry is often eliminated by inconsistent measurements from multiple viewing angles

\input{figs/nav_table}

\subsection{Path Planning Evaluation}

Furthermore, we perform navigation experiments on generated maps to evaluate the applicability of our proposed method for downstream robotics tasks. For each map, we uniformly sample 200 pairs of valid start and end points in cells unoccupied across all maps, and use A* planner with Euclidean distance heuristics on 8-connected grid to generate paths on both radar and LiDAR maps. We inflate the maps by \SI{20}{\cm} during planning to na\"{i}vely simulate the configuration space of our sensor platform, as is common practice in navigation frameworks. If the generated path collides with an occupied cell in the validation map, the path is considered a failed path. We evaluate our outcomes by the success rate of paths from radar map validated on lidar map, and those from LiDAR maps validated on radar maps (RoL and LoR for shorthand respectively).

The results are summarized in Table~\ref{tbl:nav_exp_res}. The proposed method achieves a higher success rate almost across the board, and we argue that this is because the proposed method can produce important environment geometry more consistently than baseline methods (further illustrated in Fig~\ref{fig:sequence-example-pic}). We show a set of example paths in Fig~\ref{fig:path_overlay}, where baseline methods suffer from noisy points and elimination of previously observed occupied cells, and thereby leading to paths through unnavigable space. Our proposed method maintained environment geometry while not being overly conservative and introducing spurious occupied cells in the free space, leading to a map more suitable for navigation.


\subsection{Reduce Antenna Numbers}
\input{figs/ndevice}

One important factor affecting our proposed radar mapping performance is the number of antennas used to create the synthetic aperture. The more antennas we use, the denser our coverage of the synthetic aperture becomes, resulting in a higher signal-to-noise ratio for the true target on the map. However, using more antennas requires additional computation to create the map and more physical space for the radar sensor, since antennas must be placed at least half a wavelength apart. To understand the effect of antenna number on performance, we tested the mapping performance using reduced subsets of the antennas on our radar sensor.

For our TI Cascade mmWave Radar sensor, we have 4 devices on one radar sensor, and each device controls a set of 4 receivers and 3 transmitters. We compare the mapping performance of using different numbers of devices by showing the improvement over CD, HD, and F-score in Figure~\ref{fig: n_device}. The numbers of devices from 1 to 4 correspond to a total antenna pairs of $12, 48, 108, 192$. Our results show that the improvement is minor for using more than 3 devices. This is because the antenna layout on the sensor has naturally spanned an aperture, and our vehicle is only moving at a normal walking pace. Three devices of antenna pairs are close enough to cover the synthetic aperture and reach our mapping performance limit. 
\vspace{-3mm}

%

%% file: figs/sensor_platform.tex
\begin{figure}
    \centering
    \includegraphics[width=.3\textwidth]{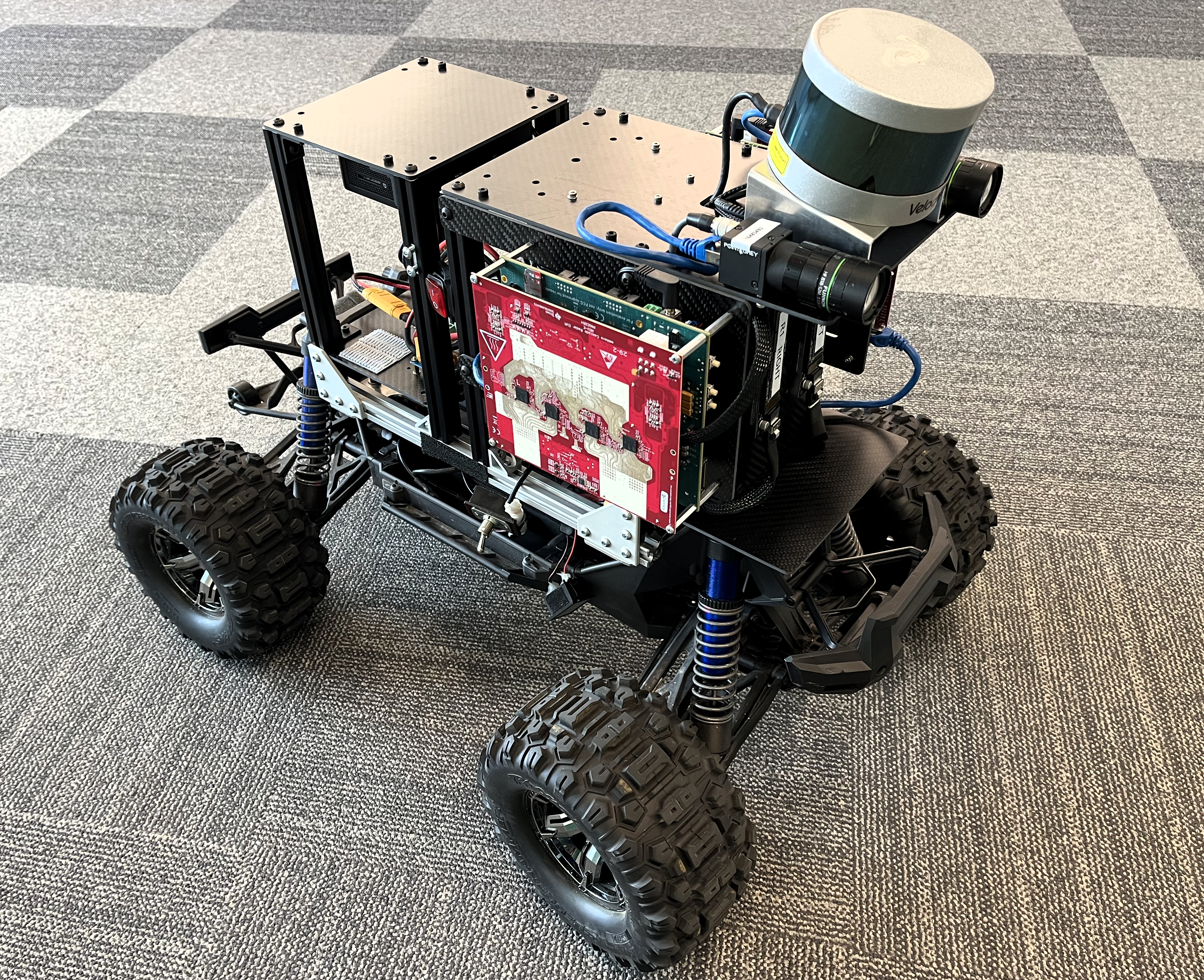}
    \caption{The ground vehicle and sensors used for our experiments. Two cascade mmWave radars are installed on both sides of the vehicle.}
    \label{fig:sensor-platform-img}
\end{figure}

%% file: figs/signal_config.tex
\begin{table}[]
\caption{Signal Configuration for Cascade Radar Sensor}
\label{tbl:radar-signal-config}

\begin{tabular}{c cccc}
  & $F_{\text{start}}$ & $F_{\text{slope}}$ & \# of RX & \# of TX \\ \hline
 Config & \SI{77}{\giga\hertz} & \SI{79}{\mega\hertz/\micro\second}  & 16       & 12   \\ [2mm]                  
  & chirp / Tx / frame & samples/chirp & chirp time &  \\ \hline
  Config & 64  & 256 & \SI{40}{\micro\second} & 
\end{tabular}

\vspace{-5mm}
\end{table}

%% file: figs/parameters_table.tex
    \begin{table}[h]                                               \vspace{-1em}
    \centering
    \caption{Significant Parameters}                                                                                    
    \begin{tabular}{lcc}
        \hline                                               
        \textbf{Parameter} & \textbf{Value} & \textbf{Description} \\
        \hline
        $\sigma_{a}$ & $0.15$ & Scale parameter of Rayleigh distribution \\
        $p_{\text{hit}}$      & $0.7$  & Upper probability clamp for occupancy update \\
        $p_{\text{miss}}$     & $0.2$  & Lower probability clamp for occupancy update \\
        $l_{\max}$            & $3.5$  & Upper map log-odds clamp (${\approx}0.97$ prob.) \\
        $l_{\min}$            & $-2.0$ & Lower map log-odds clamp (${\approx}0.12$ prob.) \\
        \hline
        
    \end{tabular}

    \label{table: parameter}
\end{table}

%% file: figs/sigma_ablation.tex
\begin{figure}
    \centering
    \includegraphics[width=0.8\linewidth]{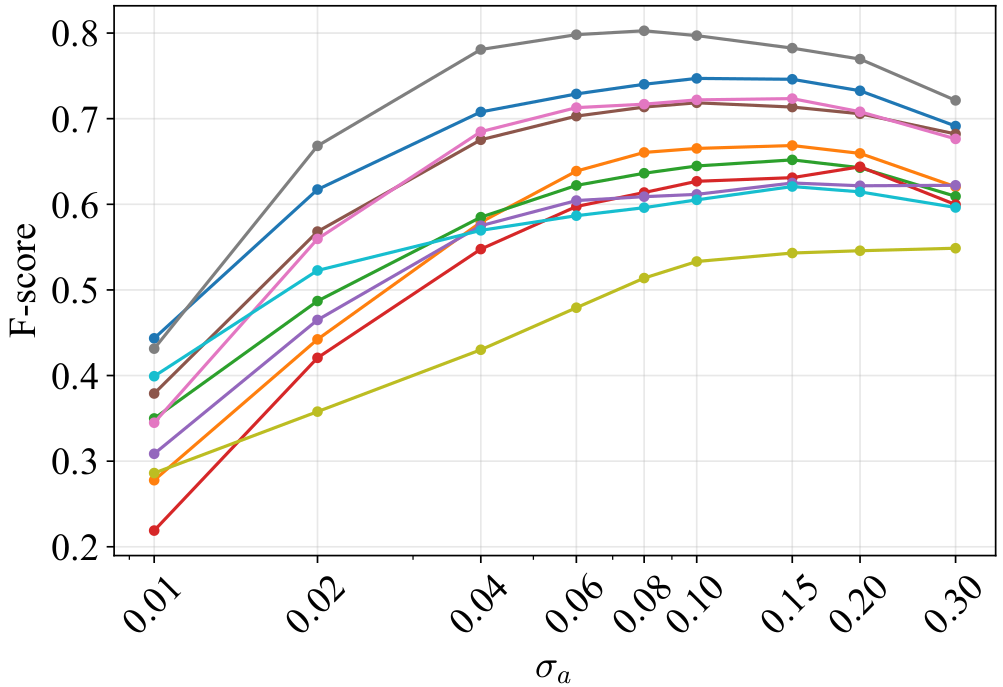}
    \caption{F-score versus the Rayleigh scale parameter $\sigma_a$ (log scale) across 10 evaluation sequences. Performance peaks around $\sigma_a = 0.15$ on average, which we adopt as the default value.}
    \label{fig: sigmas}
    \vspace{-6mm}
\end{figure}

%% file: figs/sequence_example_pic.tex
\begin{figure*}
    \centering
    
    \includegraphics[width=\linewidth]{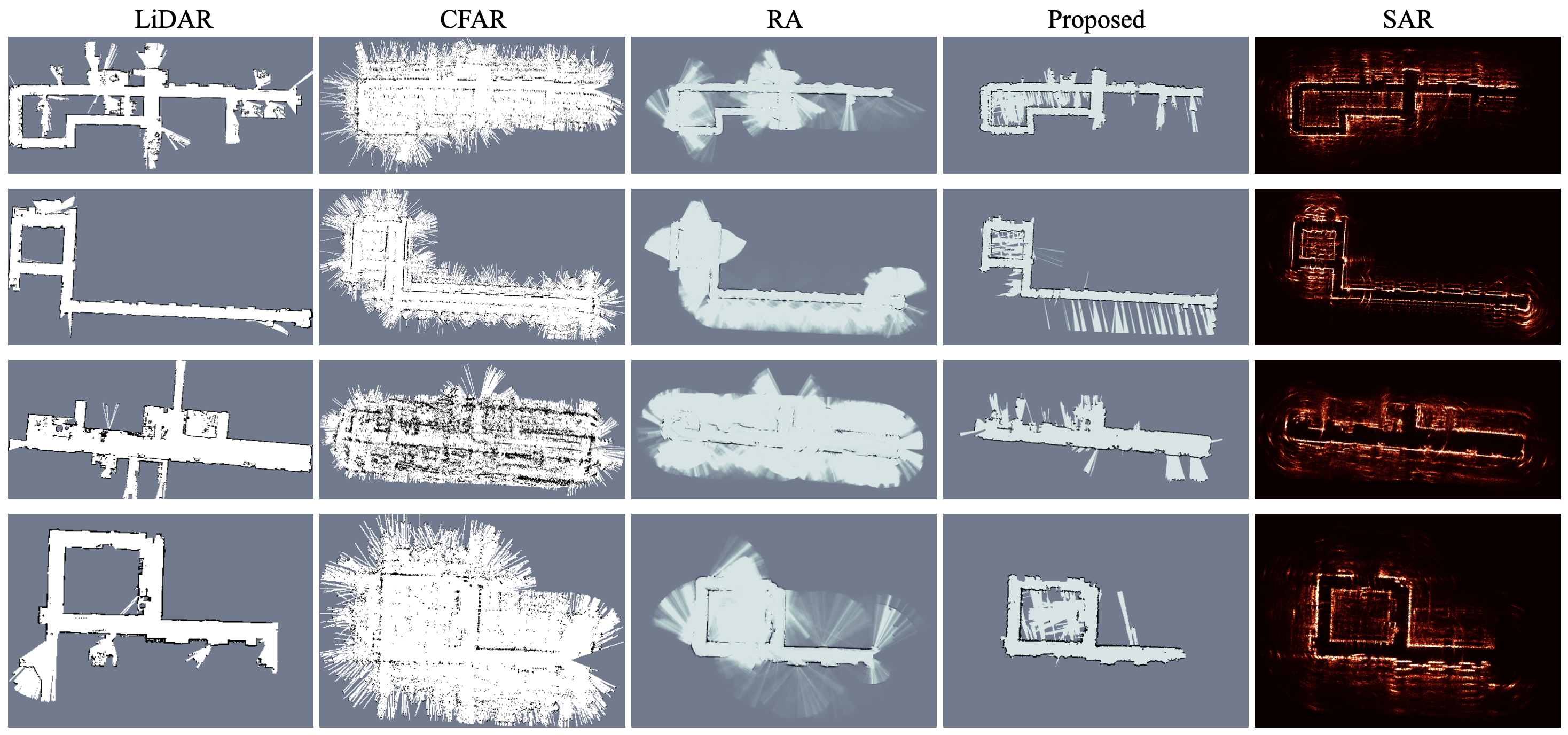}
    \caption{Example probabilistic mapping outcomes of sequences used in experiments. Our proposed approach generated maps that clearly outlined the environment and distinguished between free space and unknown space.}

    \label{fig:sequence-example-pic}
    \vspace{-0.2cm}
\end{figure*}

%% file: figs/map_quant_table.tex
\begin{table*}
    \centering
    \caption{Obstacle Evaluation Results}
    \label{tab:method_comparison}
    \resizebox{\textwidth}{!}{%
    \begin{tabular}{|l|ccc|ccc|ccc|ccc|ccc|}
    \hline
    & \multicolumn{3}{c|}{\nsh} & \multicolumn{3}{c|}{\nshshort} & \multicolumn{3}{c|}{\nsha} & \multicolumn{3}{c|}{\nshb} & \multicolumn{3}{c|}{\textbf{apart}} \\
    \textbf{Method} & CD$\downarrow$ & HD$\downarrow$ & F-score$\uparrow$ & CD$\downarrow$ & HD$\downarrow$ & F-score$\uparrow$ & CD$\downarrow$ & HD$\downarrow$ & F-score$\uparrow$ & CD$\downarrow$ & HD$\downarrow$ & F-score$\uparrow$ & CD$\downarrow$ & HD$\downarrow$ & F-score$\uparrow$ \\
    \hline
    CFAR & 1.6 & 8.605 & 0.382 & 2.92 & 9.265 & 0.289 & 2.307 & 8.908 & 0.288 & 1.694 & 8.614 & 0.38 & 2.197 & \textbf{8.8} & 0.328 \\
    RA & 1.388 & 8.414 & 0.47 & 1.239 & 8.107 & 0.574 & 0.724 & 9.119 & \textbf{0.65} & 1.422 & 8.507 & 0.357 & 0.972 & 8.957 & 0.653 \\
    Proposed & \textbf{0.853} & \textbf{7.811} & \textbf{0.642} & \textbf{0.87} & \textbf{7.912} & \textbf{0.617} & \textbf{0.6} & \textbf{6.726} & 0.633 & \textbf{1.069} & \textbf{7.043} & \textbf{0.603} & \textbf{0.609} & 9.032 & \textbf{0.731} \\
    \hline
    \hline
    & \multicolumn{3}{c|}{\textbf{square1}} & \multicolumn{3}{c|}{\textbf{square2}} & \multicolumn{3}{c|}{\textbf{z\_shape1}} & \multicolumn{3}{c|}{\textbf{z\_shape2}} & \multicolumn{3}{c|}{\wean} \\
    \textbf{Method} & CD$\downarrow$ & HD$\downarrow$ & F-score$\uparrow$ & CD$\downarrow$ & HD$\downarrow$ & F-score$\uparrow$ & CD$\downarrow$ & HD$\downarrow$ & F-score$\uparrow$ & CD$\downarrow$ & HD$\downarrow$ & F-score$\uparrow$ & CD$\downarrow$ & HD$\downarrow$ & F-score$\uparrow$ \\
    \hline
    CFAR & 2.418 & 9.013 & 0.345 & 2.317 & 8.927 & 0.397 & 2.61 & 9.604 & 0.269 & 2.793 & 8.7 & 0.268 & 2.277 & 8.8 & 0.392 \\
    RA & 0.642 & \textbf{6.276} & \textbf{0.784} & 0.923 & 7.811 & \textbf{0.764} & 1.531 & 7.743 & \textbf{0.592} & 1.664 & \textbf{9.542} & \textbf{0.604} & 1.176 & \textbf{4.952} & 0.596 \\
    Proposed & \textbf{0.638} & 6.277 & 0.707 & \textbf{0.71} & \textbf{5.1} & 0.71 & \textbf{1.26} & \textbf{4.864} & 0.496 & \textbf{1.442} & 9.934 & 0.594 & \textbf{0.519} & 5.486 & \textbf{0.799} \\
    \hline
    \end{tabular}%
    }
    \label{table: point eval}
\vspace{-0.4cm}
\end{table*}

%% file: figs/nav_table.tex

\definecolor{sarcolor}{RGB}{27,158,119}
\definecolor{racolor}{RGB}{217,95,2}
\definecolor{cfarcolor}{RGB}{169,166,208}

\DeclareRobustCommand{\sarline}{\tikz[baseline=-0.3ex]{\draw[sarcolor, line width=1.2pt, solid] (0,0) -- (0.3cm,0);}}
\DeclareRobustCommand{\raline}{\tikz[baseline=-0.3ex]{\draw[racolor, line width=1.2pt, dashed] (0,0) -- (0.3cm,0);}}
\DeclareRobustCommand{\cfarline}{\tikz[baseline=-0.3ex]{\draw[cfarcolor, line width=1.2pt, dashdotted] (0,0) -- (0.4cm,0);}}

\begin{table}[t]
\centering
\caption{Navigation Experiment Results}
\label{tbl:nav_exp_res}

\resizebox{0.9\columnwidth}{!}{%
\begin{tabular}{|c|cc|cc|cc|}
\hline
         & \multicolumn{2}{c|}{\nsh} & \multicolumn{2}{c|}{\nshshort} & \multicolumn{2}{c|}{\nsha} \\
\textbf{Method} & RoL$\uparrow$ & LoR$\uparrow$ & RoL$\uparrow$ & LoR$\uparrow$ & RoL$\uparrow$ & LoR$\uparrow$ \\ \hline
CFAR     & 0.235 & 0.275 & 0.410 & 0.475 & 0.205 & 0.095 \\
RA       & 0.345 & 0.300 & 0.640 & 0.590 & 0.430 & 0.250 \\
Proposed & \textbf{0.690} & \textbf{0.365} & \textbf{0.960} & \textbf{0.460} & \textbf{0.550} & \textbf{0.350} \\ \hline\hline
         & \multicolumn{2}{c|}{\nshb} & \multicolumn{2}{c|}{\textbf{apart}} & \multicolumn{2}{c|}{\textbf{square1}} \\
\textbf{Method} & RoL$\uparrow$ & LoR$\uparrow$ & RoL$\uparrow$ & LoR$\uparrow$ & RoL$\uparrow$ & LoR$\uparrow$ \\ \hline
CFAR     & 0.635 & 0.040 & 0.180 & 0.075 & 0.500 & 0.265 \\
RA       & 0.420 & 0.440 & 0.290 & 0.170 & 0.675 & \textbf{0.425} \\
Proposed & \textbf{0.905} & \textbf{0.655} & \textbf{0.925} & \textbf{0.280} & \textbf{0.990} & 0.425 \\ \hline
\end{tabular}
}

\vspace{2pt} 

\resizebox{0.62\columnwidth}{!}{%
\begin{tabular}{|c|cc|cc|}
\hline
         & \multicolumn{2}{c|}{\textbf{square2}} & \multicolumn{2}{c|}{\textbf{zshape\_1}} \\
\textbf{Method} & RoL$\uparrow$ & LoR$\uparrow$ & RoL$\uparrow$ & LoR$\uparrow$ \\ \hline
CFAR     & 0.360 & 0.375 & 0.205 & 0.425 \\
RA       & 0.665 & 0.295 & 0.480 & \textbf{0.350} \\
Proposed & \textbf{0.880} & \textbf{0.205} & \textbf{0.980} & 0.350 \\ \hline\hline
         & \multicolumn{2}{c|}{\textbf{zshape\_2}} & \multicolumn{2}{c|}{\wean} \\
\textbf{Method} & RoL$\uparrow$ & LoR$\uparrow$ & RoL$\uparrow$ & LoR$\uparrow$ \\ \hline
CFAR     & 0.235 & 0.300 & 0.325 & 0.325 \\
RA       & \textbf{0.950} & \textbf{0.420} & 0.475 & 0.595 \\
Proposed & 0.950 & 0.460 & \textbf{0.950} & \textbf{0.270} \\ \hline
\end{tabular}
}
\vspace{-0.5cm}
\end{table}

\begin{figure}[t]
  \centering
  \begin{minipage}[b]{0.40\columnwidth}
    \centering
    \includegraphics[width=\linewidth]{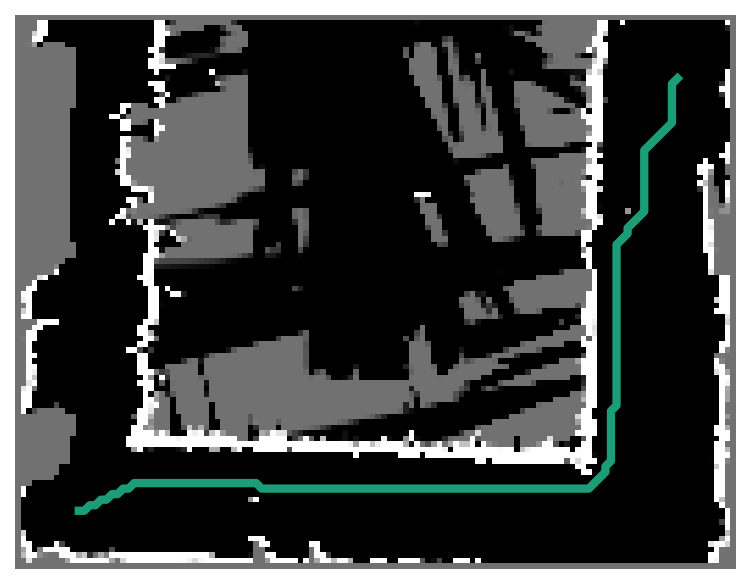}
  \end{minipage}
  \begin{minipage}[b]{0.40\columnwidth}
    \centering
    \includegraphics[width=\linewidth]{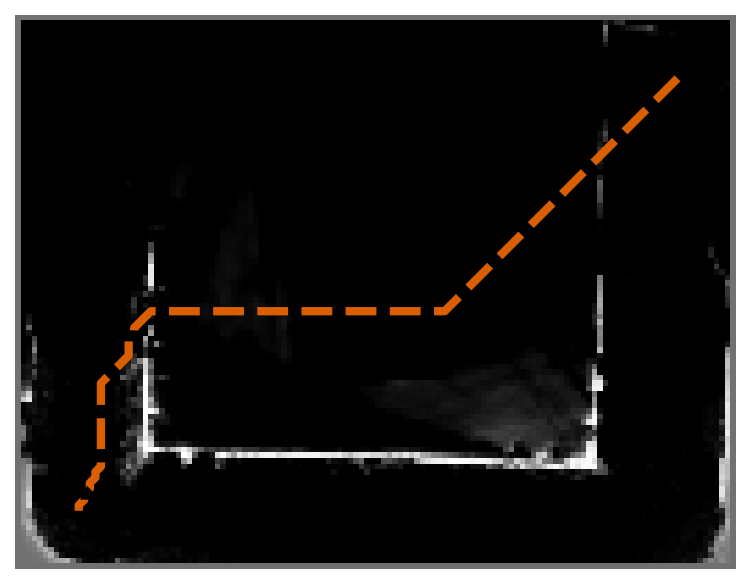}
  \end{minipage}

  \begin{minipage}[b]{0.40\columnwidth}
    \centering
    \includegraphics[width=\linewidth]{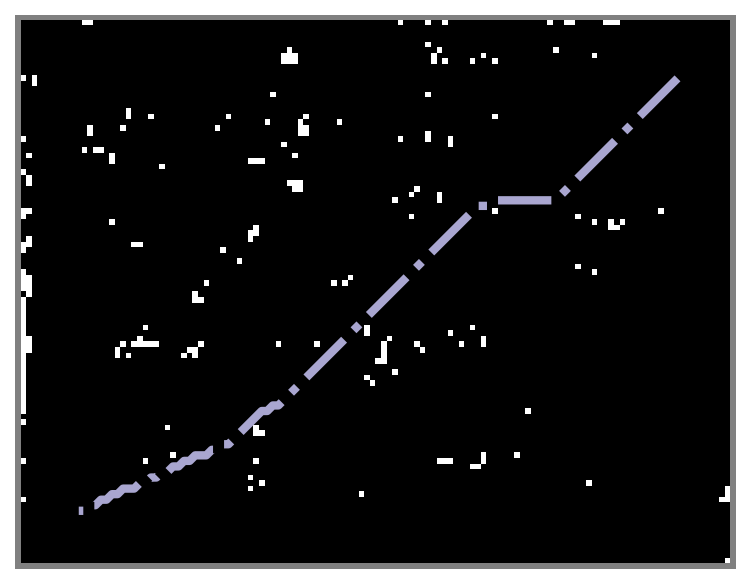}
  \end{minipage}
  \begin{minipage}[b]{0.40\columnwidth}
    \centering
    \includegraphics[width=\linewidth]{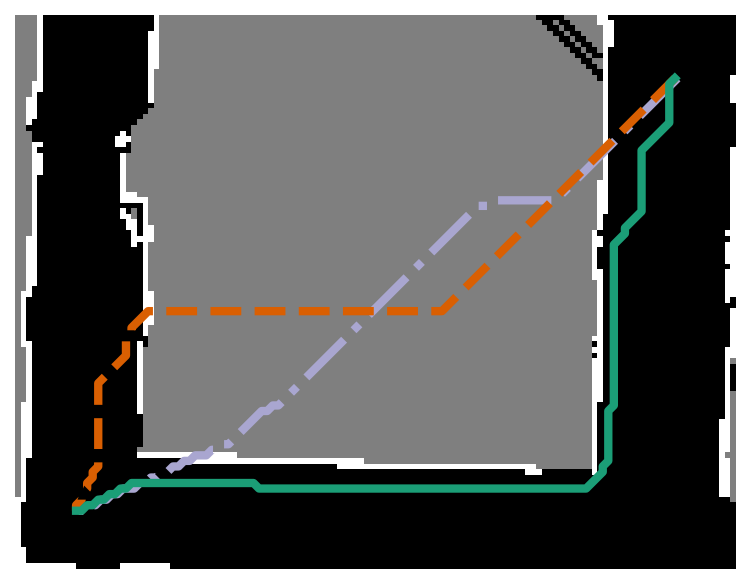}
  \end{minipage}
  \caption{Example path overlaid on their respective maps and LiDAR map. Proposed (top left, \sarline), RA (top right, \raline), CFAR (bottom left, \cfarline), and LiDAR Map (bottom right). Grey: unknown; Black: navigable; White: obstacles.}
  \label{fig:path_overlay}
  \vspace{-0.5cm}
\end{figure}

%% file: figs/ndevice.tex
\begin{figure*}
    \centering
    \includegraphics[width=\linewidth]{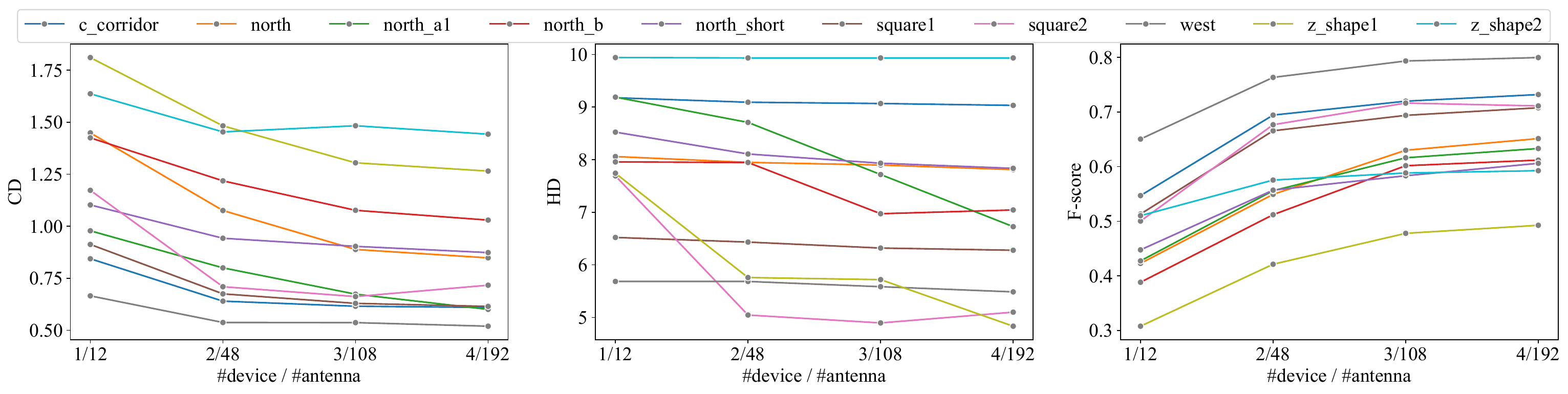}
    \caption{Mapping quality as a function of the number of antennas used. Each device contributes an additional 4 receivers and 3 transmitters.}
    \label{fig: n_device}
    \vspace{-5mm}
\end{figure*}

%% file: 05_conclusion.tex
\section{Conclusion}
In this work, we propose an approach that adapts synthetic aperture radar mapping techniques to build occupancy maps from FMCW mmWave radar signals. We compare our results across different signal processing steps and probability modeling approaches for occupancy map construction. We evaluate our results against LiDAR-generated ground truth maps and test the performance of downstream path planning tasks on the generated maps, achieving superior performance. An ablation study on the number of antennas required for map construction suggests directions for future work on reducing computational requirements for real-time robotics mapping tasks. For further improvement, efforts could focus on adjusting signal modulation parameters, testing alternative probability modeling approaches, and implementing more precise timing and localization systems to create more accurate maps. Finally, our code and data used in this paper are publicly available.

\vspace{-2mm}

%% file: ref.bib
@string{IJRR  = "Intl. J. of Robotics Research (IJRR)"}

@string{RAL   = "IEEE Robotics and Automation Letters (RA-L)"}

@string{CVPR = "Proc. IEEE Conf. on Computer Vision and Pattern Recognition (CVPR)"}

@string{WACV = "Proc. IEEE/CVF Winter Conference on Applications of Computer Vision (WACV)"}

@string{ICCV = "Proc. Intl. Conf. on Computer Vision (ICCV)"}

@string{ICRA = "Proc. IEEE Intl. Conf. on Robotics and Automation (ICRA)"}

@string{IROS = "Proc. IEEE/RSJ Intl. Conf. on Intelligent Robots and Systems (IROS)"}

@string{FUSION = "Proc. Intl. Conf. on Information Fusion (FUSION)"}

@inproceedings{kramer2020radar,
    title={Radar-inertial ego-velocity estimation for visually degraded environments},
    author={Kramer, Andrew and Stahoviak, Carl and Santamaria-Navarro, Angel and Agha-Mohammadi, Ali-Akbar and Heckman, Christoffer},
    booktitle=ICRA,
    pages={5739--5746},
    year={2020},
    month=May,
    address={Paris, {FR}}
}

@article{kramer2022coloradar, 
    title={{ColoRadar}: The direct 3D millimeter wave radar dataset}, 
    author={Kramer, Andrew and Harlow, Kyle and Williams, Christopher and Heckman, Christoffer}, 
    journal=IJRR, 
    volume={41}, 
    number={4}, 
    pages={351--360}, 
    year={2022}, 
    publisher={SAGE Publications Sage UK: London, England} 
}

@article{park20213d,
    title={3{D} ego-Motion Estimation Using low-Cost mm{W}ave Radars via Radar Velocity Factor for Pose-Graph {SLAM}},
    author={Park, Yeong Sang and Shin, Young-Sik and Kim, Joowan and Kim, Ayoung},
    journal=RAL,
    volume={6},
    number={4},
    pages={7691--7698},
    year={2021},
}

@article{doer2021x,
    author = {Doer, Christopher and Trommer, Gert F.},
    year = {2022},
    month = {02},
    pages = {329-339},
    title = {{x-RIO}: Radar Inertial Odometry with Multiple Radar Sensors and Yaw Aiding},
    volume = {12},
    journal = {Gyroscopy and Navigation}
}

@inproceedings{lu2020milliego,
    title={{milliEgo}: single-chip mm{W}ave radar aided egomotion estimation via deep sensor fusion},
    author={Lu, Chris Xiaoxuan and Saputra, Muhamad Risqi U and Zhao, Peijun and Almalioglu, Yasin and De Gusmao, Pedro PB and Chen, Changhao and Sun, Ke and Trigoni, Niki and Markham, Andrew},
    booktitle={Proc. ACM Conf. on Embedded Networked Sensor Systems},
    pages={109--122},
    year={2020},
    month=Nov,
    address={Yokohama, JP}
}

@inproceedings{wang2021rodnet,
    author={Wang, Yizhou and Jiang, Zhongyu and Gao, Xiangyu and Hwang, Jenq-Neng and Xing, Guanbin and Liu, Hui},
    title={{RODNet}: Radar Object Detection Using Cross-Modal Supervision},
    booktitle=WACV,
    month=Jan,
    year={2021},
    pages={504-513},
    address={Waikoloa, {USA}}
}

@inproceedings{wang2021rod2021,
    title={{ROD2021} Challenge: A Summary for Radar Object Detection Challenge for Autonomous Driving Applications},
    author={Wang, Yizhou and Hwang, Jenq-Neng and Wang, Gaoang and Liu, Hui and Kim, Kwang-Ju and Hsu, Hung-Min and Cai, Jiarui and Zhang, Haotian and Jiang, Zhongyu and Gu, Renshu},
    booktitle={Proc. ACM Intl. Conf. on Multimedia Retrieval (ICMR)},
    pages={553--559},
    year={2021},
    month=Nov,
    address={Taipei, {TW}}
}

@article{huang2021cross,
    title={Cross-Modal Contrastive Learning of Representations for Navigation Using Lightweight, Low-Cost Millimeter Wave Radar for Adverse Environmental Conditions},
    author={Huang, Jui-Te and Lu, Chen-Lung and Chang, Po-Kai and Huang, Ching-I and Hsu, Chao-Chun and Huang, Po-Jui and Wang, Hsueh-Cheng and others},
    journal=RAL,
    volume={6},
    number={2},
    pages={3333--3340},
    year={2021},
}

@Conference{Geneva2020ICRA,
    title = {{OpenVINS}: A Research Platform for Visual-Inertial Estimation},
    author = {Patrick Geneva and Kevin Eckenhoff and Woosik Lee and Yulin Yang and Guoquan Huang},
    booktitle = ICRA,
    Year = {2020},
    month=May,
    Address = {Paris, {FR}},
    pages={4666-4672},
}

@article{texas2020imaging,
    title={Imaging radar using cascaded mmwave sensor reference design},
    author={Texas Instruments, Inc},
    year={2020}
}

@ARTICLE{hornung13auro,
  author = {Armin Hornung and Kai M. Wurm and Maren Bennewitz and Cyrill
  Stachniss and Wolfram Burgard},
  title = {{OctoMap}: An Efficient Probabilistic {3D} Mapping Framework Based
  on Octrees},
  journal = {Autonomous Robots},
  year = 2013,
  doi = {10.1007/s10514-012-9321-0},
  note = {Software available at \url{https://octomap.github.io}}
}

@inproceedings{huang2024multi,
  title={Multi-radar inertial odometry for 3d state estimation using {mmWave} imaging radar},
  author={Huang, Jui-Te and Xu, Ruoyang and Hinduja, Akshay and Kaess, Michael},
  booktitle=ICRA,
  pages={12006--12012},
  month=may,
  year={2024}, 
  address={{Yokohama}, {Japan}}
}

@inproceedings{huang2024dart,
  title={DART: Implicit doppler tomography for radar novel view synthesis},
  author={Huang, Tianshu and Miller, John and Prabhakara, Akarsh and Jin, Tao and Laroia, Tarana and Kolter, Zico and Rowe, Anthony},
  booktitle=CVPR,
  pages={24118--24129},
  year={2024}
}

@inproceedings{panoradar,
  title={Enabling Visual Recognition at Radio Frequency},
  author={Lai, Haowen and Luo, Gaoxiang and Liu, Yifei and Zhao, Mingmin},
  booktitle={Proceedings of the 30th Annual International Conference on Mobile Computing and Networking (MobiCom)},
  pages={388--403},
  year={2024}
}

@inproceedings{wang2024vision,
  title={Vision meets {mmWave} radar: 3d object perception benchmark for autonomous driving},
  author={Wang, Yizhou and Cheng, Jen-Hao and Huang, Jui-Te and Kuan, Sheng-Yao and Fu, Qiqian and Ni, Chiming and Hao, Shengyu and Wang, Gaoang and Xing, Guanbin and Liu, Hui and others},
  booktitle={2024 IEEE Intelligent Vehicles Symposium (IV)},
  pages={2769--2775},
  year={2024},
}

@inproceedings{Xu22iros,
   author = {R. Xu and W. Dong and A. Sharma and M. Kaess},
   title = {Learned Depth Estimation of {3D} Imaging Radar for Indoor Mapping},
   booktitle = IROS,
   address = {Kyoto, Japan},
   month = oct,
   year = {2022}
}

@article{zhang2024towards,
  title={Towards dense and accurate radar perception via efficient cross-modal diffusion model},
  author={Zhang, Ruibin and Xue, Donglai and Wang, Yuhan and Geng, Ruixu and Gao, Fei},
  journal=RAL,
  year={2024},
}

@inproceedings{prabhakara2023high,
  title={High resolution point clouds from mmwave radar},
  author={Prabhakara, Akarsh and Jin, Tao and Das, Arnav and Bhatt, Gantavya and Kumari, Lilly and Soltanaghai, Elahe and Bilmes, Jeff and Kumar, Swarun and Rowe, Anthony},
  booktitle=ICRA,
  pages={4135--4142},
  year={2023},
  month=jun,
  address={London, {UK}},
}

@article{ding2024radarocc,
  title={RadarOcc: Robust 3D occupancy prediction with 4D imaging radar},
  author={Ding, Fangqiang and Wen, Xiangyu and Zhu, Yunzhou and Li, Yiming and Lu, Chris Xiaoxuan},
  journal={Advances in Neural Information Processing Systems},
  volume={37},
  pages={101589--101617},
  year={2024}
}

@inproceedings{mopidevi2024rmap,
  title={RMap: Millimeter-wave radar mapping through volumetric upsampling},
  author={Mopidevi, Ajay Narasimha and Harlow, Kyle and Heckman, Christoffer},
  booktitle=IROS,
  pages={1108--1115},
  year={2024},
  month=oct,
  address={abudhabi, {UAE}}
}

@inproceedings{lu2020see,
  title={See through smoke: robust indoor mapping with low-cost mmwave radar},
  author={Lu, Chris Xiaoxuan and Rosa, Stefano and Zhao, Peijun and Wang, Bing and Chen, Changhao and Stankovic, John A and Trigoni, Niki and Markham, Andrew},
  booktitle={Proc. of the 18th International Conference on Mobile Systems, Applications, and Services},
  pages={14--27},
  year={2020}
}

@inproceedings{wang2018high,
  title={High-resolution image synthesis and semantic manipulation with conditional {GANs}},
  author={Wang, Ting-Chun and Liu, Ming-Yu and Zhu, Jun-Yan and Tao, Andrew and Kautz, Jan and Catanzaro, Bryan},
  booktitle=CVPR,
  pages={8798--8807},
  year={2018}
}

@inproceedings{borts2024radar,
  title={Radar fields: Frequency-space neural scene representations for fmcw radar},
  author={Borts, David and Liang, Erich and Broedermann, Tim and Ramazzina, Andrea and Walz, Stefanie and Palladin, Edoardo and Sun, Jipeng and Brueggemann, David and Sakaridis, Christos and Van Gool, Luc and others},
  booktitle={ACM SIGGRAPH 2024 Conference Papers},
  pages={1--10},
}

@article{yanik2020development,
  title={Development and demonstration of MIMO-SAR mmWave imaging testbeds},
  author={Yanik, Muhammet Emin and Wang, Dan and Torlak, Murat},
  journal={IEEE Access},
  volume={8},
  pages={126019--126038},
  year={2020},
}

@inproceedings{iqbal2021realistic,
  title={Realistic SAR implementation for automotive applications},
  author={Iqbal, Hasan and L{\"o}ffler, Andreas and Mejdoub, Mohamed Nour and Gruson, Frank},
  booktitle={IEEE 17th European Radar Conference (EuRAD)},
  pages={306--309},
  year={2021},
}

@article{grebner2022radar,
  title={Radar-based mapping of the environment: Occupancy grid-map versus SAR},
  author={Grebner, Timo and Schoeder, Pirmin and Janoudi, Vinzenz and Waldschmidt, Christian},
  journal={IEEE Microwave and Wireless Components Letters},
  volume={32},
  number={3},
  pages={253--256},
  year={2022},
}

@article{grebner2023probabilistic,
  title={Probabilistic SAR processing for high-resolution mapping using millimeter-wave radar sensors},
  author={Grebner, Timo and Grathwohl, Alexander and Schoeder, Pirmin and Janoudi, Vinzenz and Waldschmidt, Christian},
  journal={IEEE Transactions on Aerospace and Electronic Systems},
  volume={59},
  number={5},
  pages={4800--4814},
  year={2023},
}

@inproceedings{ritterbusch2024indoor,
  title={Indoor synthetic aperture radar measurements of point-like targets using a wheeled mobile robot},
  author={Ritterbusch, Yuma E and Fink, Johannes and Waldschmidt, Christian},
  booktitle={EUSAR 2024; 15th European Conference on Synthetic Aperture Radar},
  pages={595--600},
  year={2024},
  organization={VDE}
}

@article{ritterbusch2024rio,
  title={RIO-SAR: Synthetic aperture radar imaging of indoor scenes based on radar-inertial odometry using a mobile robot},
  author={Ritterbusch, Yuma E and Fink, Johannes and Waldschmidt, Christian},
  journal={IEEE Transactions on Radar Systems},
  year={2024},
}

@article{ritterbusch2024simultaneous,
  title={Simultaneous Localization and Mapping for Indoor Mobile Robots using Synthetic Aperture Radar Images},
  author={Ritterbusch, Yuma Elia and Fink, Johannes and Waldschmidt, Christian},
  journal={IEEE Journal of Indoor and Seamless Positioning and Navigation},
  year={2024},
}

@inproceedings{werber2015automotive,
  title={Automotive radar gridmap representations},
  author={Werber, Klaudius and Rapp, Matthias and Klappstein, Jens and Hahn, Markus and Dickmann, J{\"u}rgen and Dietmayer, Klaus and Waldschmidt, Christian},
  booktitle={IEEE MTT-S International Conference on Microwaves for Intelligent Mobility (ICMIM)},
  pages={1--4},
  year={2015},
}

@inproceedings{prophet2018adaptions,
  title={Adaptions for automotive radar based occupancy gridmaps},
  author={Prophet, Robert and Stark, Henriette and Hoffmann, Marcel and Sturm, Christian and Vossiek, Martin},
  booktitle={IEEE MTT-S International Conference on Microwaves for Intelligent Mobility (ICMIM)},
  pages={1--4},
  year={2018},
}

@inproceedings{weishaupt2022precfar,
  title={Pre{CFAR} gridmaps for automotive radar},
  author={Weishaupt, Fabio and Appenrodt, Nils and Tilly, Julius F and Dickmann, J{\"u}rgen and Heberling, Dirk},
  booktitle={18th European Radar Conference (EuRAD)},
  pages={161--164},
  year={2022},
  organization={IEEE}
}

@inproceedings{degerman20163d,
  title={3D occupancy grid mapping using statistical radar models},
  author={Degerman, Johan and Pernst{\aa}l, Thomas and Alenljung, Klas},
  booktitle={IEEE Intelligent Vehicles Symposium (IV)},
  pages={902--908},
  year={2016},
}

@article{frey2009focusing,
  title={Focusing of airborne synthetic aperture radar data from highly nonlinear flight tracks},
  author={Frey, Othmar and Magnard, Christophe and Ruegg, Maurice and Meier, Erich},
  journal={IEEE transactions on geoscience and remote sensing},
  volume={47},
  number={6},
  pages={1844--1858},
  year={2009},
  publisher={IEEE}
}

@article{ulander2003synthetic,
  title={Synthetic-aperture radar processing using fast factorized back-projection},
  author={Ulander, Lars MH and Hellsten, Hans and Stenstrom, Gunnar},
  journal={IEEE Transactions on Aerospace and electronic systems},
  volume={39},
  number={3},
  pages={760--776},
  year={2003},
  publisher={IEEE}
}

@book{curlander1991synthetic,
  title={Synthetic aperture radar},
  author={Curlander, John C and McDonough, Robert N},
  volume={11},
  year={1991},
  publisher={Wiley, New York}
}

@article{kuruoglu2004modeling,
  title={Modeling SAR images with a generalization of the Rayleigh distribution},
  author={Kuruoglu, Ercan E and Zerubia, Josiane},
  journal={IEEE Transactions on Image Processing},
  volume={13},
  number={4},
  pages={527--533},
  year={2004},
}

@article{xu2021fast,
  title={Fast-lio: A fast, robust lidar-inertial odometry package by tightly-coupled iterated kalman filter},
  author={Xu, Wei and Zhang, Fu},
  journal={IEEE Robotics and Automation Letters},
  volume={6},
  number={2},
  pages={3317--3324},
  year={2021},
}

@article{kung2025radarsplat,
  title={RadarSplat: Radar Gaussian Splatting for High-Fidelity Data Synthesis and 3D Reconstruction of Autonomous Driving Scenes},
  author={Kung, Pou-Chun and Harisha, Skanda and Vasudevan, Ram and Eid, Aline and Skinner, Katherine A},
  journal={arXiv preprint arXiv:2506.01379},
  year={2025}
}

@InProceedings{Huang_2025_ICCV,
    author    = {Huang, Tianshu and Prabhakara, Akarsh and Chen, Chuhan and Karhade, Jay and Ramanan, Deva and O'toole, Matthew and Rowe, Anthony},
    title     = {Towards Foundational Models for Single-Chip Radar},
    booktitle = {Proceedings of the IEEE/CVF International Conference on Computer Vision (ICCV)},
    month     = {October},
    year      = {2025},
    pages     = {24655-24665}
}
